\documentclass{bmvc2k}

\usepackage{wrapfig}
\usepackage[ruled,vlined,linesnumbered]{algorithm2e}
\usepackage{algorithmic}
\usepackage{dsfont}
\usepackage{amsmath}
\usepackage{amssymb}
\usepackage{pifont}
\usepackage{booktabs} 
\usepackage{cleveref}
\usepackage{graphicx}
\usepackage{multicol}
\usepackage{multirow}

\crefname{table}{Tab.}{Tabs.}
\Crefname{table}{Tab.}{Tabs.}
\crefname{figure}{Fig.}{Figs.}
\Crefname{figure}{Fig.}{Figs.}
\crefname{equation}{Eq.}{Eqs.}
\Crefname{equation}{Eq.}{Eqs.}
\crefname{section}{Sec.}{Secs.}
\Crefname{section}{Sec.}{Secs.}

\title{PPS: Plug-and-Play Saccadic Vision \\for Fine-Grained Classification}

\addauthor{Johann Schmidt}{johann.schmidt@ovgu.de}{1}
\addauthor{Sebastian Stober}{stober@ovgu.de}{1}
\addauthor{Joachim Denzler}{joachim.denzler@uni-jena.de}{2}
\addauthor{Paul Bodesheim}{paul.bodesheim@uni-jena.de}{2}

\addinstitution{
 Otto von Guericke University\\
 Universitaetsplatz 2\\
 39106 Magdeburg\\
 Germany
}
\addinstitution{
  Friedrich Schiller University\\
  Fuerstengraben 1\\
  07737 Jena\\
  Germany
}

\runninghead{Schmidt, Stober, Denzler, Bodesheim}{Plug-and-Play Saccadic Vision}

\begin{document}

\maketitle

\begin{abstract}
Fine-grained visual classification benefits from high-resolution inputs to resolve subtle, localized cues, but at the cost of large compute and memory budgets. 
Inspired by human saccadic vision, we propose a lightweight extension that turns any pretrained backbone with spatial feature maps into a saccadic classifier through glimpse-induced inference. 
The only architectural additions are a small priority head and a single cross-attention fusion block.
The same weight-tied encoder encodes a downsampled peripheral glance, exposes the spatial feature map from which the priority head predicts a glimpse-priority distribution and re-encodes high-resolution glimpse crops sampled from that distribution in parallel.
A sequential sampler with non-maximum suppression draws a runtime-configurable number of glimpses.
A residual multi-head cross-attention block then fuses the glance against the glimpses. 
To improve the quality of the priority maps, we train the priority head by a per-glimpse information-gain target from a lightweight per-image memory bank. 
This requires no reinforcement learning.
Complexity is linear in the number of glimpses and peak memory is decoupled from input resolution. 
Across nine benchmarks our framework improves consistently over its backbones.
\end{abstract}
\begin{figure}[h!]
    \centering
    \includegraphics[width=0.7\linewidth]{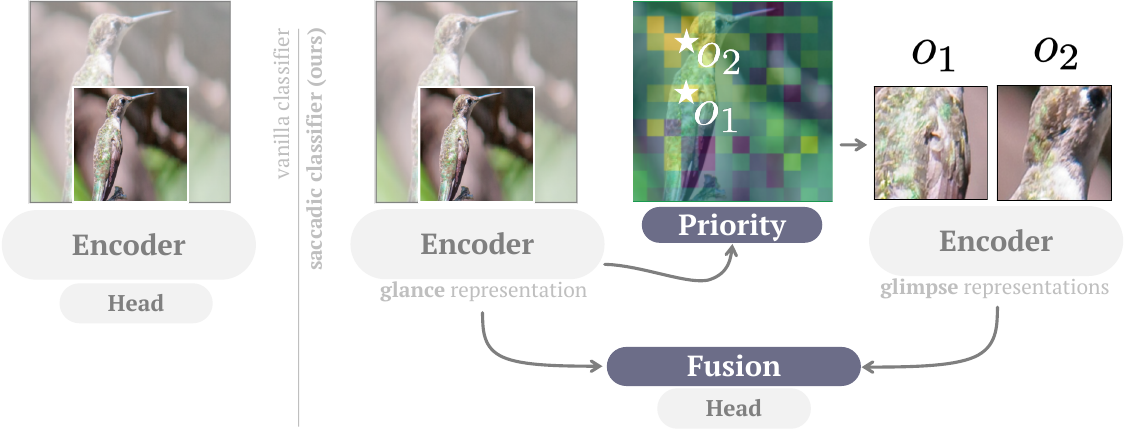}
    \caption{
    We reuse the backbone encoder to capture a glance, generate a priority map and encode sampled glimpses. 
    Glance and glimpse representations are fused for prediction.
    }
    \label{fig:starter}
\end{figure}

\section{Introduction}
\label{sec:intro}

Fine-grained visual classification (FGVC) has emerged as a critical area of research within computer vision, aiming to distinguish between visually very similar categories, such as different bird species \cite{CUB200, NABirds} or insect breeds \cite{EU, Ecuador, AMI}.
Unlike traditional image classification, FGVC focuses on identifying subtle, local patterns and nuanced variations across categories, which is crucial for applications like biodiversity monitoring \cite{EU}.
Despite its significance, FGVC remains challenging due to inherent complexities:
(i) data is often sparse, (ii) categories exhibit high intra-class variability (pose, lighting, occlusion), (iii) inter-class differences are typically subtle and localized.
While (i) and (ii) are tackled by large pre-trained backbones, (iii) requires models to process high-resolution signals to capture minute visual details.
To keep the emerging costs low, model reuse is essential.
Recent methods take inspiration from human visual perception, which is highly efficient and effective.
Instead of uniformly scanning a scene, the human visual system employs rapid eye movements, known as \emph{saccades} \cite{Yarbus1967}, interspersed with brief, high-acuity \emph{glimpses}  (fixations) \cite{Itti2001}.
This visual search process is not random but guided by various sources of pre-attentive information that coalesce to form a spatial \emph{priority map} \cite{Wolfe2021, Itti2001}.
This map dynamically directs selective attention to the most salient locations.
Once attention is focused, the high-acuity central vision (fovea) extracts fine-grained details, while the rest of the scene is processed via peripheral vision at a lower resolution (the \emph{glance}) \cite{Javier2010}.
This allows for capturing a high-resolution image using multiple low-resolution windows \cite{Ballard1991}.
In computer vision, this strategy is realized by glimpse based models, which form a prediction from a few localized views instead of a glance alone \cite{RAM, RACNN, GFNet, Saccader}. 
Because the prediction rests on the extracted views, these models stay interpretable and faithful \cite{PartCNN, MPSA}.
Yet three design choices recur across them and limit reuse. 
Most attach a dedicated localization network to the recognition backbone \cite{RACNN, GFNet, NTSNet}.
The glimpse count is fixed at architecture time \cite{RACNN, DPS, NTSNet}.
And placing or aggregating views relies on reinforcement learning (RL) or sequential encoding \cite{RAM, Saccader, GFNet}. 

\paragraph{Our Contributions}
We introduce the \emph{Plug-and-Play Saccader} (PPS)\footnote{
Our source code is available at \texttt{github.com/johSchm/Saccader}.
}, a saccadic extension that grants any vision classifier high resolution glimpses at a cost set by the glance resolution and a small glimpse budget rather than the input resolution.
One weight tied encoder encodes a low resolution peripheral glance, reuses the resulting feature map as a spatial priority map, and encodes the sampled glimpse patches. 
This removes the separate localization network and the memory overhead of prior glimpse models \cite{RACNN, GFNet, Saccader}. 
Our contributions are:
\begin{itemize}
    \item
    PPS wraps any backbone that emits spatial feature maps, convolutional or transformer, including those built on sparse token subsets. 
    The shared encoder processes the glance and all glimpses alike, so input resolution and glimpse count become runtime settings rather than architectural commitments.
    \item
    Glimpses are read directly from the encoder's own feature map and supervised by an auxiliary priority loss, leaving training fully end to end without RL.
    \item
    A single cross-attention block aggregates all glimpses and fuses them with the glance in one pass. 
    Glimpses are thus processed in parallel rather than sequentially. 
    Across nine FGVC benchmarks and three backbones, PPS improves consistently over the vanilla backbone. 
\end{itemize}

\section{Related Work}
\label{sec:relatedwork}

\begin{figure}
    \centering
    \includegraphics[width=1.0\linewidth]{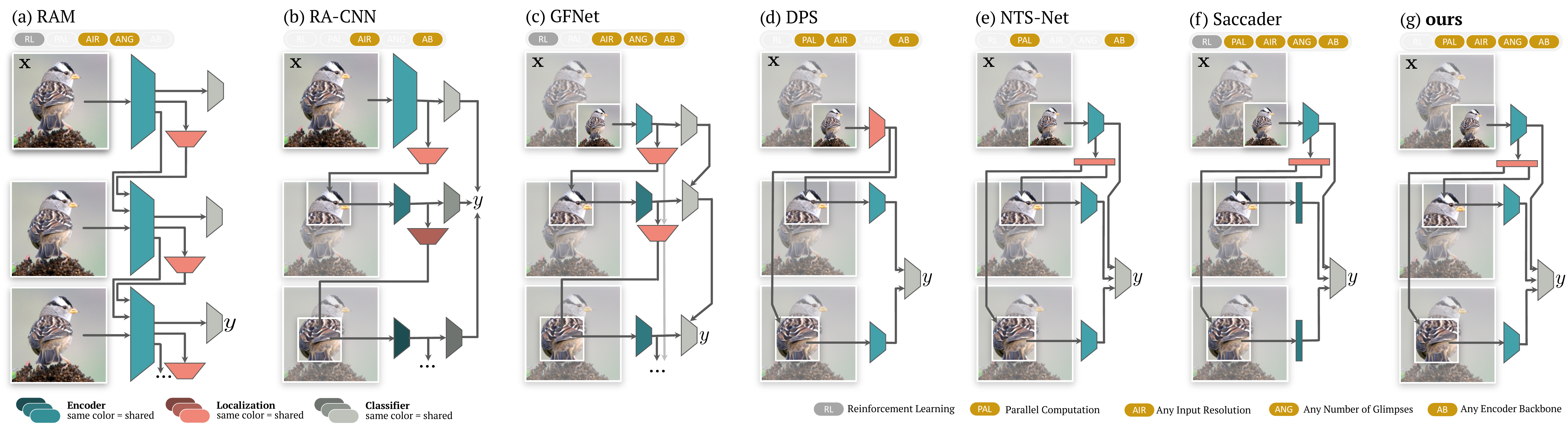}
    \caption{
    Visual side-by-side comparison of related glimpse-based approaches covering RAM \cite{RAM}, RA-CNN \cite{RACNN}, GFNet \cite{GFNet}, DPS \cite{DPS}, NTS-Net \cite{NTSNet}, Saccader \cite{Saccader} and our saccadic vision extension.
    The sharing of encoders, additional localization networks for glimpse localization, and the classifier are highlighted by different colors.
    Tags are used to highlight training based on reinforcement learning, whether glimpse encoding is parallelized (or sequential) and whether the approach is agnostic to the input resolution, the number of glimpses or the choice of backbone.
    }
    \label{fig:saccader_related_work}
\end{figure}

Glimpse-based vision models extend a classifier with a mechanism that selects a small set of localized views and aggregates evidence from them, rather than processing the image uniformly.
RAM \cite{RAM} casts this as a reinforcement-learning problem in which a recurrent controller emits glimpse locations and is rewarded by the final classification accuracy.
RA-CNN \cite{RACNN} replaces the controller with a stack of attention proposal networks that recursively zoom into a single discriminative region, NTS-Net \cite{NTSNet} introduces a separate navigator whose proposals are ranked by a teacher classifier, and Saccader \cite{Saccader} returns to an explicitly saccadic policy trained with reinforcement learning.
GFNet \cite{GFNet} and DPS \cite{DPS} reduce compute by selecting glimpses from coarse feature maps, but still attach a dedicated localization or routing network on top of the recognition backbone.
As summarized in \Cref{fig:saccader_related_work}, the recurring design commitments across this body of work are an additional localization network, a fixed number of glimpses, and either reinforcement learning or sequential glimpse encoding.

Our framework removes these commitments: glimpse priorities are read directly from the encoder's own feature map with a single weight-tied backbone, all glimpses are processed in parallel, the input resolution and the number of fixations are runtime-configurable, and training is fully end-to-end without reinforcement learning.
Within FGVC more broadly, part-based feature localization can be grouped into two complementary regimes: \emph{Single-Patch Part} (SPP) methods, which extract rectangular crops from the source image, and \emph{Multi-Patch Part} (MPP) methods, which extract sparse token subsets from an encoded feature map.
Our approach falls into the SPP family but is compatible with both convolutional and transformer backbones (including MPP encoders).

\paragraph{SPP methods.}
SPP methods have their roots in object detection \cite{RCNN, FastRCNN, FasterRCNN, MaskRCNN, YOLOv1, SSD, YOLOv1}, where two-stage detectors process individual region proposals at high accuracy but with significant computational overhead, while single-stage detectors combine proposal and prediction in a single forward pass for improved efficiency.
In FGVC, similar techniques are used to extract object parts \cite{PartCNN}, but no ground-truth part annotations are available.
Glimpses (part crops) are therefore learned implicitly, guided by the classification loss.
To impose some structure on the extracted parts, several frameworks introduce semantically clustered parts \cite{Dongliang2021, MPSA}.
Part-Stacked CNNs \cite{PartCNN} adopt a single-stage framework to simultaneously generate region proposals and integrate their encodings for classification, while Dongliang et al.\ \cite{Dongliang2021} learn multi-grained parts using top-down spatial attention.

To locate glimpses, many frameworks take inspiration from region proposal networks \cite{FasterRCNN} and either regress locations and ratios without inductive biases \cite{RAM} or read out activation peaks on spatial location maps \cite{RACNN, Saccader, GFNet}.
Either a fixed number of glimpses are extracted \cite{RACNN, DPS, NTSNet, AFTrans, TransFG} or recurrent mechanisms sample a variable number \cite{RAM, Saccader, GFNet}.
Variable-count sampling typically requires reinforcement learning \cite{RAM, Saccader, GFNet}, which renders training challenging.
Many frameworks also rely on encoders tailored to a specific sub-task such as feature extraction for the peripheral view, for glimpses, or for locations, complicating the use of off-the-shelf backbones \cite{RAM, PartCNN}.
Because these sub-tasks extract very similar features, the resulting encoders are highly redundant; this slows down learning and increases computational demand.
This motivated per-encoder losses to ensure consistent and robust learning signals \cite{Zhang2021FGVC} and weight-tied encoders \cite{DPS, NTSNet, Xu2024}.
A critical aspect of these methods is the choice of context representation, which can rely on high-resolution images \cite{EU} or downsampled views \cite{PartCNN}; computational demand grows quickly when the entire high-resolution input must be encoded \cite{RAM, RACNN, PartCNN}.
Finally, glimpse features are fused with peripheral features through recurrent encoding \cite{RAM, GFNet}, concatenation \cite{RACNN, PartCNN}, multi-scale cross-attention between parts \cite{Xu2024}, or graph attention modelling explicit spatial relationships \cite{Sikdar2024}.

\paragraph{MPP methods.}
MPP methods encode the image as a whole and interpret the resulting feature maps as hierarchies of part maps, potentially sparsified by thresholds.
While this view applies to ConvNets, it is most prominent in vision transformers \cite{ViT, DeiT}, where the global attention mechanism naturally yields sparse part maps.
To remove background noise, improve efficiency, and further sparsify the representation, token sparsification (token dropout or pruning) is commonly used; this process corresponds to localization in SPP methods.
TransFG \cite{TransFG} uses accumulated attention scores for token dropout.
DynamicViT \cite{DynamicViT} performs dynamic token pruning that creates spatially coherent token selections, which inherently form part maps.
A-ViT \cite{A-ViT} introduces a halting probability that enables adaptive token selection, where high-probability tokens correspond to discriminative parts.
ACC-ViT \cite{ACC-ViT} mixes high-attention tokens from different classes, effectively isolating class-specific part regions and excluding background noise.
Evo-ViT \cite{EvoViT} uses a slow--fast token evolution mechanism that dynamically separates informative from redundant tokens to accelerate inference.
FAL-ViT \cite{FAL-ViT} applies coarse-to-fine token masking to create hierarchical part representations, progressing from global object structure to fine-grained details.
PIM \cite{PIM} uses a graph neural network to model intra- and inter-stage relationships between the top-$k$ preserved tokens.
HERBS \cite{HERBS} retains only top-$k$ spatial features at different network stages, with higher temperatures in early layers to encourage exploration while providing guidance.

\section{Methodology}
\label{sec:method}

\begin{figure}
    \centering
    \includegraphics[width=1.0\linewidth]{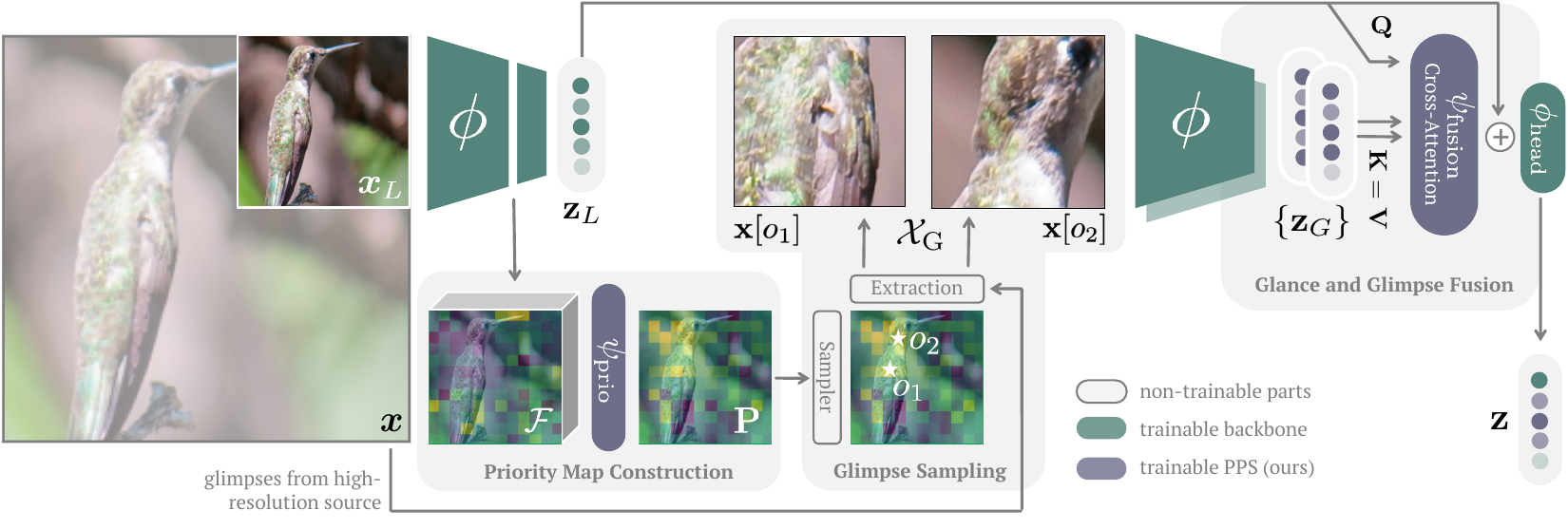}
    \caption{
    \textbf{Overview of PPS.}
    A glance $\mathbf{x}_L$ is obtained by downsampling the input $\mathbf{x}$.
    The backbone $\phi$ encodes it into spatial feature maps $\mathcal{F}$ and a glance embedding $\mathbf{z}_L$.
    A lightweight priority head $\psi_{\text{prio}}$ turns $\mathcal{F}$ into a priority map $\mathbf{P}$, read as a spatial distribution over locations.
    From $\mathbf{P}$ we sample $N$ glimpse locations $\mathcal{O}$ and crop the corresponding patches $\mathcal{X}_G$ from the high resolution source $\mathbf{x}$.
    The same (weight-tied) backbone $\phi$ encodes each patch into a glimpse embedding, giving $\{\mathbf{z}_G\}$.
    A cross-attention block $\psi_{\text{fusion}}$ fuses the glimpse set with the glance $\mathbf{z}_L$, and the head $\phi_{\text{head}}$ maps the result to class logits $\mathbf{z}$.
    }
    \label{fig:method}
\end{figure}

Let $(\mathbf{x}, y) \sim \mathcal{D}$ represent a sampled tuple from a dataset $\mathcal{D}$, where $\mathbf{x} \in \mathbb{R}^{C \times H \times W}$ is an image and $y \in \mathbb{N}$ is its corresponding class label.
$C \in \mathbb{N} \setminus \{ 0 \}$ is the number of channels (we use $C=3$ for RGB images) and $H,W \in \mathbb{N} \setminus \{ 0 \}$ denote the spatial resolution of the pixel grid.
To obtain a glance, $\mathbf{x}$ is downsampled to $\mathbf{x}_L \in \mathbb{R}^{C \times H_L \times W_L}$ with $H_L \ll H$ and $W_L \ll W$.
Let $(\phi, \phi_{\text{head}})$ be a vision backbone with encoder $\phi: \mathbb{R}^{C \times H_L \times W_L} \to \mathbb{R}^D$ and head $\phi_{\text{head}}: \mathbb{R}^{D} \to \mathbb{R}^Y$, where $D \in \mathbb{N} \setminus \{ 0 \}$ is embedding dimension and $Y \in \mathbb{N} \setminus \{ 0 \}$ is the number of labels.
Let $\mathcal{F} = \{\mathbf{F}^{(s)}\}_{s=1}^{S}$ denote the encoder's spatial feature maps over $S$ encoding stages on the glance view $\mathbf{x}_{L}$. 
Our proposed framework equips any vision backbone with the ability to leverage higher resolution inputs with only very little additional parameters.
The process is illustrated in \Cref{fig:method} and discussed in the following.

\subsection{Constructing Priority Maps}
\label{sec:priomap_construction}

SPP and MPP methods rely on a map that marks where to look, or which tokens deserve attention. 
A common approach trains a dedicated proposal network \citep{RACNN, GFNet}, sometimes conditioned on the raw input \citep{DPS}.
These networks are large and must learn the mapping from scratch. Following \cite{NTSNet, Saccader}, we reuse the backbone's own representations and learn only a shallow head on top. 
The intuition: features used to \emph{recognize} an object already encode where to \emph{look}. 
We retain only the final map $\mathbf{F}^{(S)} \in \mathbb{R}^{C^{(S)} \times H_P \times W_P} \in \mathcal{F}$, whose channels carry the most discriminative responses. 
Aggregating one head per stage $s \in [1,\ldots,S]$ did not improve accuracy and computation costs. 
A shallow convolutional head $\psi_{\text{prio}}: \mathbb{R}^{C^{(S)} \times H_P \times W_P} \to \mathbb{R}^{H_P \times W_P}$ projects $\mathbf{F}^{(S)}$ to a scalar priority map $\mathbf{P} \in \mathbb{R}^{H_P \times W_P}$. 
This generalizes the fixed Gaussian smoothing of \citet{PartCNN} to a learned channel reweighting. 
The head adds negligible parameters, since $\mathbf{F}^{(S)}$ is already computed by the underlying vision backbone.
As $\mathbf{F}^{(S)}$ is only trained for classification purposes and $\mathbf{P}$ is conditioned on it, $\mathbf{P}$ shows no accumulation at borders (sign of spatial collapses).
This removes the need for an additional edge mask as in \citet{GFNet}.
We support our claims empirically in \Cref{sec:experiments}.

\subsection{Sampling Glimpses from the Priority Map}
\label{sec:sampling}

\begin{figure}[t]
    \centering
    \begin{minipage}[t]{0.19\textwidth}
        \vspace{0pt}
        \includegraphics[width=\linewidth]{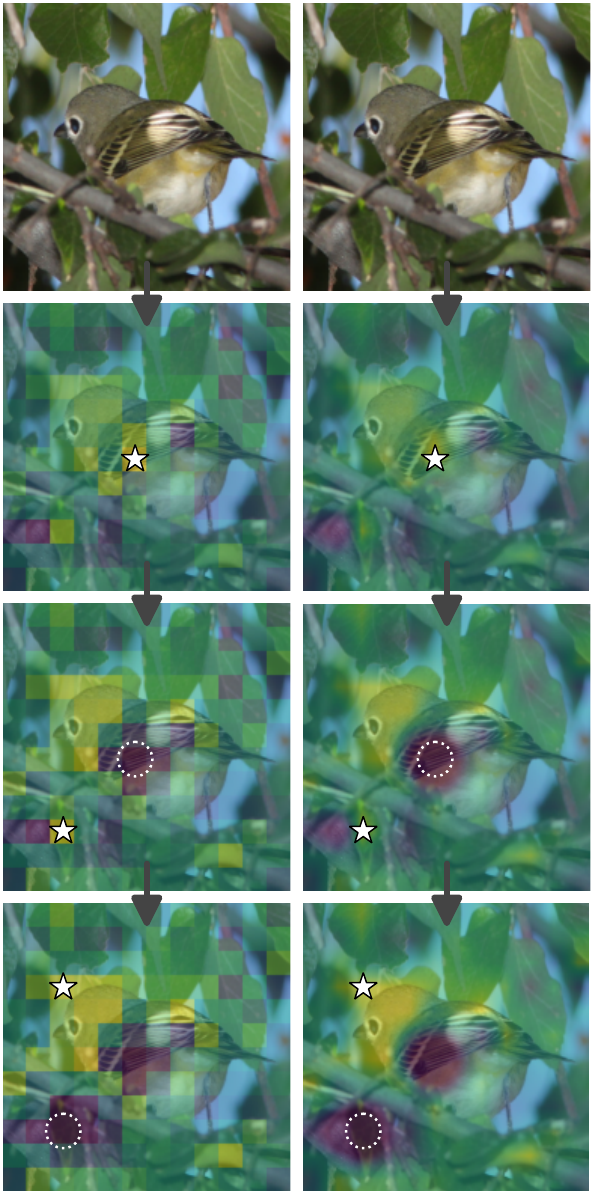}
    \end{minipage}%
    \hfill
    \begin{minipage}[t]{0.75\textwidth}
        \vspace{0pt}
    \begin{algorithmic}[1]
        \REQUIRE prior map $\mathbf{P}$, number of glances $N$
        \ENSURE  glimpse locations $\mathcal{O} \subset [0,1]^2$ with $|\mathcal{O}| = N$
        \STATE \textbf{Init:} $\mathcal{O} \leftarrow \emptyset$
        \STATE $\Omega_{ij} \leftarrow \bigl((j+\tfrac{1}{2})/W,\ (i+\tfrac{1}{2})/H\bigr)$ \COMMENT{normal. centered grid}
        \FOR{$n = 1$ to $N$} 
            \STATE $k \leftarrow \arg\max_{H_P W_P} \left( \operatorname{vec} \left( \mathbf{P} \right) \right)$ 
            \STATE $\mathbf{o} \leftarrow \Omega_{\lfloor k/W \rfloor,\ k \bmod W}$ \COMMENT{index position on 2D grid}
            \STATE $K_{ij} \leftarrow \exp\!\bigl(-\|\Omega_{ij}-\mathbf{o}\|_2^2 / (2\sigma^2)\bigr)$ \COMMENT{std. deviation $\sigma$}
            \STATE $\mathbf{P} \leftarrow \mathbf{P} - \lambda K$ \COMMENT{additive $\lambda$-weighted penalty}
            \STATE $\mathcal{O} \leftarrow \mathcal{O} \cup \{\mathbf{o}\}$ \COMMENT{add the estimated glimpse location}
        \ENDFOR
        \RETURN $\mathcal{O}$
    \end{algorithmic}
    \end{minipage}
    \vspace{0.2cm}
    \caption{
    The priority progression over $N$ glimpses using our NMS with Gaussian kernels on $\mathbf{P}$.
    Left an example progression for two different resolutions of $\mathbf{P}$ and on the right the pseudo-code of our sampling method.}
    \label{alg:suppression}
    \vspace{-0.3cm}
\end{figure}

\noindent
Direct coordinate regression \cite{STN}, as used in prior work \cite{GFNet, RACNN, Saccader}, does not encode spatial structure and fails to generalize \cite{Nibali2018}.
We instead exploit the spatial structure of $\mathbf{P}$ and draw $N$ glimpse locations $\mathcal{O} = \{\mathbf{o} \mid \mathbf{o} \sim_N \mathbf{P}\}$ from it.
Unlike prior work \cite{RACNN, NTSNet, GFNet} that fixes the number of glimpses architecturally, our sampler treats $N$ as a runtime hyperparameter.
Sampling budgets thus change without retraining or any architectural modification.
We denote the train- and test-time counts by $N_{\text{train}}, N_{\text{test}} \in \mathbb{N}$.
Setting $N=0$ recovers the vanilla backbone forward pass, so our PPS can be disabled whenever walltime outweighs accuracy.

Locations are expressed in normalized, centered coordinates $2\mathbf{o}-1 \in [-1,1]^2$, decoupling regression from the input resolution $H \times W$.
Pixel coordinates follow as $[(\mathbf{o}+1)/2] \odot [H, W]^\top$.
This lets us estimate locations directly on the low-resolution feature map $\mathbf{F} \in \mathbb{R}^{C \times H_P \times W_P}$ with $H_P, W_P \ll H, W$, at its native resolution.
No upsampling is required: the gradient signal stays concentrated on each cell of $\mathbf{F}$ rather than being diffused across pixels.
Continuous, pixel-precise localization \cite{DPS} is unnecessary for FGVC and only adds compute (see \Cref{sec:ablation}).
A coarse prior map also stabilizes training, as the smaller discrete location set reduces variance in the sampler's output.
To avoid redundant samples in adjacent regions, we apply non-maximum suppression (NMS) during sampling.
Rather than discrete IoU-based suppression on detections \cite{NTSNet}, we apply sequential Gaussian soft-suppression on $\mathbf{P}$.
A smooth radial kernel masks already-selected neighborhoods, with no anchors or box parameters required.
The only hyperparameters are the kernel variance (we use $\sigma=0.04$) and the update weight $\lambda$ (we use $\lambda=1e^{-4}$).
Locations are selected greedily by \texttt{argmax} \cite{NTSNet, DPS}.
\Cref{alg:suppression} gives the complete procedure with an example of suppression progression.
Given locations $\mathcal{O}$, we extract a fixed-size glimpse of resolution $H_L \times W_L$ at each by a differentiable affine crop \cite{STN}.
Sampling uses bicubic interpolation with (zero) border padding, yielding crops of identical resolution independent of the source size $H \times W$.
This decouples the glimpse resolution from the input resolution. 
The result are a set of glimpses $\mathcal{X}_G = \{\mathbf{x}^{n}\}_{n=1}^{N}$, which are encoded next.

\subsection{Fusing Glance and Glimpse Representations}
\label{sec:fusion}

We reuse the same backbone encoder $\phi: \mathbb{R}^{C \times H_L \times W_L} \to \mathbb{R}^{D}$, where $D$ denotes the embedding dimension, for the glance and glimpse encoding.
This is applied to each glimpse in $\mathcal{X}_G$, producing a set of glimpse embeddings $\{\mathbf{z}_{G}^{n}\}_{n=1}^{N}$ with $\mathbf{z}_{G}^{n} \in \mathbb{R}^{D}$.
Furthermore, we have the glance embedding $\mathbf{z}_{L} \in \mathbb{R}^{D}$.
The goal is to update the glance features by the glimpse features by fusion of these embeddings vectors.
The informativeness of the two pathways is strongly instance-dependent.
Some target objects fill most of the canvas, so that the localized glimpses are largely redundant with the glance context.
Others are small and off-center, so that the discriminative evidence resides in a few glimpses while the remaining ones capture uninformative background.
A fusion rule with a fixed \cite{Saccader} or globally predicted mixing weight cannot represent this.
We therefore replace static weighting by a \emph{residual cross-attention} block $\psi_{\text{fusion}}: (N+1) \cdot \mathbb{R}^D \to \mathbb{R}^D$ in which the glance representation queries the glimpses, and the mixing coefficients are produced per glimpse, per attention head, and conditioned on the glance content itself.
Aggregation over the $N$ glimpses and fusion with the glance are thereby performed jointly by a single operator.
We treat the glance embedding as a single query token and the $N$ glimpse embeddings as the keys and values of a multi-head attention layer.
Let $h$ denote one of $H$ heads with $d = D/H$.
Both streams are first layer-normalized with independent statistics and linearly projected
\begin{equation}
    \mathbf{q}^{(h)} = \mathbf{W}_{Q}^{(h)}\,\operatorname{LN}_{q}(\mathbf{z}_L),
    \qquad
    \mathbf{k}_{n}^{(h)} = \mathbf{W}_{K}^{(h)}\,\operatorname{LN}_{kv}(\mathbf{z}_{G}^{n}),
    \qquad
    \mathbf{v}_{n}^{(h)} = \mathbf{W}_{V}^{(h)}\,\operatorname{LN}_{kv}(\mathbf{z}_{G}^{n}),
\end{equation}
with $\mathbf{W}_{Q}^{(h)}, \mathbf{W}_{K}^{(h)}, \mathbf{W}_{V}^{(h)} \in \mathbb{R}^{d \times D}$.
The attention distribution over glimpses for head $h$ is the scaled dot-product softmax
\begin{equation}
\label{eq:attnweights}
    a_{n}^{(h)}
    = \frac{\exp\!\big( \langle \mathbf{q}^{(h)}, \mathbf{k}_{n}^{(h)} \rangle / \sqrt{d}\, \big)}
           {\sum_{m=1}^{N} \exp\!\big( \langle \mathbf{q}^{(h)}, \mathbf{k}_{m}^{(h)} \rangle / \sqrt{d}\, \big)},
    \qquad
    a_{n}^{(h)} \ge 0,
    \quad
    \sum_{n=1}^{N} a_{n}^{(h)} = 1 .
\end{equation}
Each head returns the induced convex combination of glimpse values.
The head outputs are concatenated and linearly mixed, and a further output projection yields the glimpse summary
\begin{equation}
\label{eq:glimpsesummary}
    \mathbf{o}^{(h)} = \sum_{n=1}^{N} a_{n}^{(h)}\, \mathbf{v}_{n}^{(h)},
    \qquad
    \hat{\mathbf{z}}_{G}
    = \mathbf{W}_{g}\,\big[\,\mathbf{W}_{O}\,
      \operatorname{concat}\!\big( \mathbf{o}^{(1)}, \dots, \mathbf{o}^{(H)} \big)\big],
\end{equation}
where $\mathbf{W}_{O} \in \mathbb{R}^{D \times D}$ is the projection internal to the attention layer and $\mathbf{W}_{g} \in \mathbb{R}^{D \times D}$ is an additional learned output projection (followed, during training, by dropout on $\hat{\mathbf{z}}_{G}$).
The fused representation is formed by a residual connection that adds the glimpse summary onto the \emph{normalized} glance, which is then encoded by the linear classification head, such that
\begin{equation}
\label{eq:per_fix_fusion}
    \mathbf{z} =
    \phi_{\text{head}} \left( \operatorname{LN}_{\text{fusion}}(\hat{\mathbf{z}}_{L} + \hat{\mathbf{z}}_{G}) \right) =
    \phi_{\text{head}} \left( \operatorname{LN}_{\text{fusion}}(\operatorname{LN}_{q}(\mathbf{z}_L) + \hat{\mathbf{z}}_{G}) \right),
\end{equation}
where $\phi_{\text{head}}: \mathbb{R}^D \to \mathbb{R}^Y$ is the head of the backbone projecting to the class logits.
The residual is essential as our PPS first has to learn where to look, so glimpses are not very useful early in training.
When no glimpses are taken ($N = 0$) the fusion block is bypassed and the model reduces to a pure-glance classifier (resembling the vanilla backbone pass).


\subsection{Learning to Focus}
\label{sec:prioloss}

\begin{figure}
    \centering
    \includegraphics[width=0.8\linewidth]{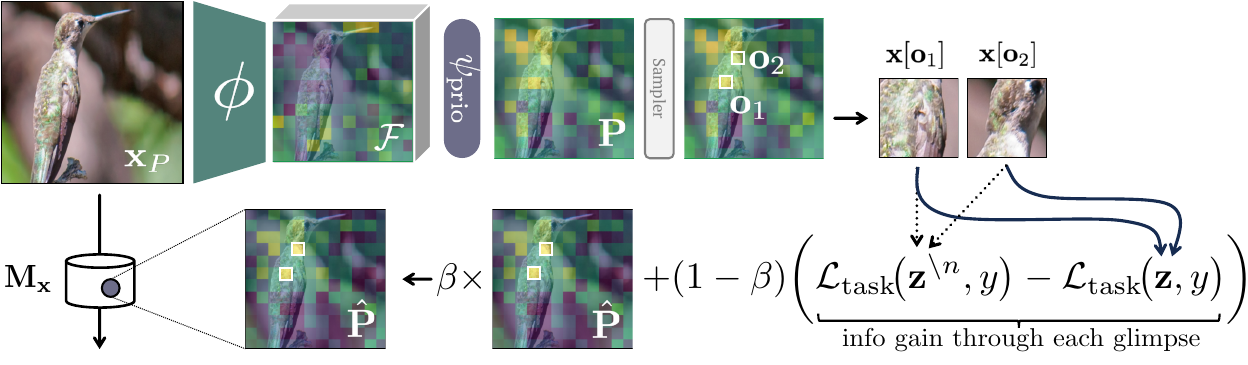}
    \caption{The training process of PPS with focus on our auxiliary priority loss.}
    \label{fig:prio_map_training}
\end{figure}

Our model is trained end-to-end.
The downstream task supplies a per-example loss $\mathcal{L}_{\text{task}}(\mathbf{z}, y)$ on the fused logits $\mathbf{z}$ of \Cref{eq:per_fix_fusion}.
For classification this is the categorical cross-entropy against the label $y$.
However, the \texttt{argmax} used to select glimpse locations in \Cref{sec:sampling} is non-differentiable\footnote{In \Cref{sec:ablation} we test with a Gumbel softmax, rendering PPS fully differentiable but also reducing performance.}, so no gradient of $\mathcal{L}_{\text{task}}$ reaches the priority map $\mathbf{P}$.
Prior work \cite{RAM, GFNet, Saccader} closes this gap with reinforcement learning, whose policy-gradient estimators are high-variance, slow to converge, and reliant on hand-tuned baselines.
We instead detach the sampler from the backward graph and supervise $\mathbf{P}$ with an auxiliary priority loss $\mathcal{L}_{\text{prio}}$.
The full objective is
\begin{equation}
\label{eq:total_loss}
    \mathcal{L} = \mathcal{L}_{\text{task}}(\mathbf{z}, y) + \lambda_{\text{prio}}\,\mathcal{L}_{\text{prio}},
\end{equation}
with $\lambda_{\text{prio}} > 0$ a balancing scalar, which we set to $\lambda_{\text{prio}}=1$.
The procedure is illustrated in \Cref{fig:prio_map_training} and discussed in the following.

\paragraph{Glimpse Utility.}
We quantify the contribution of each glimpse by a leave-one-out (LOO) ablation in logit space.
This is computed under exponential-moving-average (EMA) teacher parameters for the classification head and the fusion $(\bar{\theta}_{\,\text{head}}, \bar{\theta}_{\,\text{fusion}})$.
Let $\mathbf{z}^{\setminus n}$ denote the fused logits from \Cref{eq:per_fix_fusion} with glimpse $n$ masked out of the cross-attention keys and values.
The utility of glimpse $n$ is the resulting loss change,
\begin{equation}
\label{eq:loo_delta}
    \Delta \mathcal{L}_{n} = \mathcal{L}_{\text{task}}\!\big(\mathbf{z}^{\setminus n}, y\big) - \mathcal{L}_{\text{task}}\!\big(\mathbf{z}, y \big).
\end{equation}
A positive $\Delta \mathcal{L}_{n}$ marks a glimpse that carried discriminative evidence.
A near-zero value marks a redundant or background crop.
$\Delta \mathcal{L}_{n}$ is the pointwise information gain that glimpse $n$ contributes to the teacher's log-posterior over $y$, so the priority loss aligns $\mathbf{P}$ with expected per-glimpse informativeness.
All $N{+}1$ passes share the encoder embeddings and run as a single batched call on the lightweight fusion and head subnets, so the overhead is negligible.
The teacher is refreshed after each step by $\bar{\theta}_{\,\text{head}} \leftarrow m\bar{\theta}_{\,\text{head}} + (1-m)\theta_{\,\text{head}}$\footnote{Here, $m \in (0,1)$ is a EMA momentum term, which we set to $m=0.99$ for all our experiments.} (and likewise for $\bar{\theta}_{\,\text{fusion}}$), which insulates the targets from optimizer noise while tracking the evolving feature space.
Furthermore, backbones without registers, such as Swin \cite{Swin}, can emit stray high-norm activations \cite{RegisterToken}.
Anchoring $\mathbf{P}$ to measured glimpse utility rather than raw feature magnitude removes the issue without backbone-specific fixes.

\paragraph{Priority Memory Bank.}
A single step observes only $N$ of the $H_P W_P$ cells, so per-step utilities are too sparse to supervise $\mathbf{P}$ directly.
We accumulate them in a persistent bank $\mathbf{M} \in \mathbb{R}^{|\mathcal{D}_{\text{train}}| \times H_P \times W_P}$.
With $\hat{\mathbf{o}} = \lfloor [H_P, W_P] \odot \mathbf{o} \rfloor$ denoting the cell index, each visit updates the bank by
\begin{equation}
\label{eq:memory_update}
    \mathbf{M}_{\mathbf{x}}\!\left[\hat{\mathbf{o}}_n\right] \leftarrow \beta\,\mathbf{M}_{\mathbf{x}}\!\left[\hat{\mathbf{o}}_n\right] + (1-\beta)\,\Delta \mathcal{L}_{n},
\end{equation}
with $\beta \in [0,1)$, where $\beta = 0$ is used on the first visit.
$\mathbf{M}$ is initialized with zeros and updated over training time. 
Repeated visits across epochs yield a temporally smoothed utility field over the full grid.

\paragraph{Priority Loss.}
We turn the priority logits $\mathbf{P}$ and the bank $\mathbf{M}$ into distributions over the $H_P W_P$ cells and align them by Kullback--Leibler divergence,
\begin{equation}
\label{eq:sac_loss}
    \mathcal{L}_{\text{prio}} = \mathbb{E}_{\mathbf{x} \in \mathcal{D}_{\text{train}}}\!\left[\, D_{\mathrm{KL}}\!\big(\operatorname{softmax}(\operatorname{vec}\mathbf{M}) \,\big\|\, \operatorname{softmax}(\operatorname{vec}\mathbf{P})\big) \,\right],
\end{equation}
with the target $\mathbf{M}$ detached (stopped gradient).
The softmax discards the instance-dependent scale and offset of $\mathbf{M}$ and supervises only the relative ranking of locations.
A peaked target pulls $\mathbf{P}$ toward a confident, selective distribution.
$\mathbf{P}$ learns to concentrate probability mass on the locations the LOO measurement identifies as informative.
If $\mathcal{L}_{\text{prio}}$ would reach the encoder it would pronounce shortcut learning as the network reshaped its features to make the saliency easier to fit rather than learning where to look.
We therefore stop the gradient at $\mathbf{F}$ that conditions $\mathbf{P}$.\footnote{A joint backbone objective would force the encoder to recognize \emph{what} the object is and shape \emph{where} the model looks at once, which we found to destabilize training.}
In this setup, $\mathcal{L}_{\text{task}}$ trains the encoder, the fusion, and the head, while $\mathcal{L}_{\text{prio}}$ trains the priority encoder alone.
Hence, the two terms in \Cref{eq:total_loss} never compete for the same parameters.

\begin{figure}
    \centering
    \includegraphics[width=1.0\linewidth]{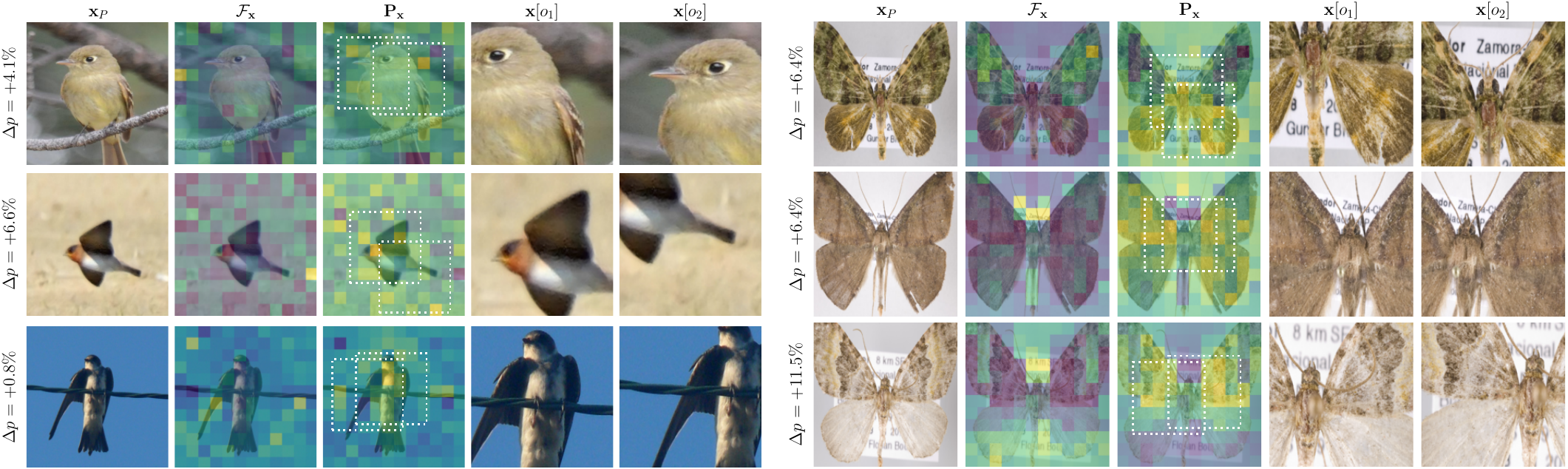}
    \caption{
    Test samples from NABirds and ECU with results obtained by our PPS-MPSA-SwinB.
    From left to right the columns show the input image $\mathbf{x}_L$, MPSA's multi-granularity feature map (generally $\mathcal{F}$), our priority map $\mathbf{P}$, and two glimpses at $\mathbf{o}_1$ and $\mathbf{o}_2$.
    Annotated on the left is the prediction delta induced by the glimpses compared to the glance.
    }
    \label{fig:examples_nabirds_ecu}
\end{figure}

\begin{table*}[t]
\centering
\caption{Average top-1 test accuracy on FGVC moth benchmarks.
}
\label{tab:moths}
\small
\begin{tabular}{@{}lccccc|llll@{}}
\toprule
\textbf{Method} & \textbf{Backbone} & \textbf{Pretr.} & \textbf{Res.} & \textbf{ECU} & \textbf{EU} & \textbf{WE} & \textbf{NA} & \textbf{CA} \\ 
\midrule
Swin-B \cite{Swin} & - & Im22k & 224 & 75.41 & 97.06 & 94.03 & 97.57 & 98.98 \\
MPSA \cite{MPSA} & SwinB & Im22k & 224 & 77.04 & 97.64 & 94.17 & 97.68 & 99.07 \\
PPS-MPSA (ours) & SwinB & Im22k & 224 & \textbf{80.89} & \textbf{98.18} & \textbf{94.44} & \textbf{97.76} & \textbf{99.08} \\
\midrule
SPPs-Cond. \cite{Korsch2022} & Inc.V3 & Im1k & 299 & - & 91.50 & - & - & - \\
SPPs-Cond. \cite{Korsch2022} & Inc.V3 & iNat & 299 & - & 93.13 & - & - & - \\
\midrule
Swin-B \cite{Swin} & - & Im22k & 384 & 77.19 & 97.10 & 94.84 & 97.95 & 99.02 \\
MPSA \cite{MPSA} & SwinB & Im22k & 384 & 80.30 & 98.36 & 94.78 & 97.94 & \textbf{99.09} \\
PPS-MPSA (ours) & SwinB & Im22k & 384 & \textbf{83.41} & \textbf{98.55} & \textbf{94.96} & \textbf{98.09} & 99.08 \\
\bottomrule
\end{tabular}%
\end{table*}

\section{Experiments}
\label{sec:experiments}


\begin{wraptable}{R}{0.55\textwidth}
\centering
\caption{Average top-1 test accuracy on CUB.}
\label{tab:fgvc-cub}
\setlength{\tabcolsep}{4pt}
\small
\begin{tabular}{@{}clccc@{}}
\toprule
 & Method & Backbone & Pretr. & Acc@1 \\
\midrule
\multirow{5}{*}{\rotatebox[origin=c]{90}{224\,px}}
  & RA-CNN \cite{RACNN} & VGG19 & Im1k & 85.3 \\
  & TASN \cite{TASN} & ResNet50 & Im1k & 87.0 \\
  & NTSN \cite{NTSNet} & ResNet50 & Im1k & 87.3 \\
  & MPSA \cite{MPSA} & SwinB & Im22k & 91.4 \\
  & PPS-MPSA (ours) & SwinB & Im22k & \textbf{91.5} \\
\cmidrule(l){2-3}
\multirow{5}{*}{\rotatebox[origin=c]{90}{384\,px}}
  & CNB \cite{ConvNext2} & - & Im22k & 90.2 \\
  & PPS-CNB (ours) & - & Im22k & 90.5 \\
  & MetaFormer2 \cite{MetaFormer} & - & Im22k & 91.8 \\
  & MPSA \cite{MPSA} & SwinB & Im22k & 92.6 \\
  & PPS-MPSA (ours) & SwinB & Im22k & \textbf{92.7} \\
\cmidrule(l){2-3}
\multirow{5}{*}{\rotatebox[origin=c]{90}{448\,px}}
  & AFTrans \cite{AFTrans} & ViT-B16 & Im1k & 91.5 \\
  & TransFG \cite{TransFG} & ViT-B16 & Im1k & 91.7 \\
  & FAL-ViT \cite{FAL-ViT} & ViT-B16 & Im1k & 91.7 \\
  & ACC-ViT \cite{ACC-ViT} & ViT-B16 & Im1k & 91.8 \\
  & FET \cite{FET} & SwinB & Im22k & 92.9 \\
\bottomrule
\vspace{-0.5cm}
\end{tabular}
\end{wraptable}

\textbf{Backbones and Pretraining}
PPS is backbone-agnostic and applies to any vision encoder that exposes a spatial feature map, including both convolutional and transformer architectures.
All reported experiments use a ConvNextv2-Base (short CNB) \cite{ConvNext2}, a Swin~Base \cite{Swin}, a MPSA encoder \cite{MPSA} (32 heads, parts ratio 2) with a Swin~Base backbone.
Our models are pretrained on ImageNet \cite{ImageNet} (iNaturalist \cite{iNaturalist} pretraining usually increase performance even more).
We finetuned the ViT, SwinB, MPSA, and FET using their code and hyperparameters on several datasets.
Even under these settings, we are not able to exactly replicate all claimed performances and hence report our achieved values here.

\begin{wraptable}{R}{0.55\textwidth}
\centering
\vspace{-0.5cm}
\caption{Avg. top-1 test accuracy on NABirds.}
\label{tab:fgvc-nabirds}
\setlength{\tabcolsep}{4pt}
\small
\begin{tabular}{@{}clccc@{}}
\toprule
 & Method & Backbone & Pretr. & Acc@1 \\
\midrule
\multirow{3}{*}{\rotatebox[origin=c]{90}{224\,px}}
  & MetaFormer2 \cite{MetaFormer} & - & Im1k & 89.7 \\
  & MPSA \cite{MPSA} & SwinB & Im22k & 90.2 \\
  & PPS-MPSA (ours) & SwinB & Im22k & \textbf{90.4} \\
\cmidrule(l){2-3}
\multirow{3}{*}{\rotatebox[origin=c]{90}{384\,px}}
  & MetaFormer2 \cite{MetaFormer} & - & Im22k & 92.2 \\
  & MPSA \cite{MPSA} & SwinB & Im22k & 92.4 \\
  & PPS-MPSA (ours) & SwinB & Im22k & \textbf{92.5} \\
\cmidrule(l){2-3}
\multirow{4}{*}{\rotatebox[origin=c]{90}{448\,px}}
  & CrossX \cite{CrossX} & ViT-B16 & Im1k & 88.9 \\
  & TransFG \cite{TransFG} & ViT-B16 & Im1k & 90.8 \\
  & FAL-ViT \cite{FAL-ViT} & ViT-B16 & Im1k & 91.1 \\
  & ACC-ViT \cite{ACC-ViT} & ViT-B16 & Im1k & 91.4 \\
\bottomrule
\end{tabular}
\end{wraptable}

\textbf{Hyperparameters}
We follow the training recipe from \cite{MPSA} with adaptations detailed.
The peripheral input is downsampled to $H_F = W_F = 7$ grid cells for MPSA and $H_F = W_F = 14$ for SwinB.
The resolution of glimpses is specified but generally $224 \times 224$ or $384 \times 384$ (specified in each experiment) using a crop scale of $25\%$.
The fusion block uses 32-head residual cross-attention.
The priority memory bank uses decay $\beta = 0.99$.
Optimization uses AdamW \cite{AdamW} except for NABirds, CUB200, and Food101 where we used plain SGD with cosine learning-rate schedule.
We use 5 warmup epochs, gradient clipping at norm 10, label smoothing 0.1 and 0.001 weight decay.
Learning rates are dataset specific and vary between $6e^{-2}$ and $8e^{-5}$, and early stopping.
If not otherwise stated we use 512px source images, which we downsample (bilinear interpolation) to the input size of the vision backbone.
We extract $2$ glimpses during training and $4$ during testing of the same input size from the high resolution source image.
This choice follows from findings in \cite{Xu2024, DPS}, which we reduced during training to decrease walltime costs.
We use standard augmentations comprising random cropping and light Gaussian blurring ($p = 0.1$).
Because the random crop offset is symmetric and zero-mean, accumulating utilities in crop-frame cells smooths each image's bank entry by at most one cell on MPSA's $7{\times}7$ grid (sub-cell in expectation) without biasing its location, and the deterministic center crop at test time matches this expected frame.
Training runs for 50--100 epochs and mini-batches depending on dataset size on A40 and V100 GPUs.
We use a frozen seed $ =42$.
For full reproducibility of our findings we make our code, configurations, and checkpoints publicly available.

\textbf{Datasets}
We evaluate PPS on moth, bird, dog, and food benchmarks, spanning a wide range of object scales and background conditions. 
This includes five moth FGVC benchmarks: EU \cite{EU} (1650 images, 200 species), ECU \cite{Ecuador} (1445 images, 675 species) with significant center bias, and AMI-GBIF \cite{AMI} subsets\footnote{Due to link corruptions in the original AMI collection, we only included classes with $\geq 10$ samples per class.} for North America (NA) (203k images, 2408 species), Central America (CA) (18k images, 551 species) and Western Europe (33k images, 2533 species) with a significantly higher variability and noise rate. 
EU and ECU feature relatively clean backgrounds, ideal for evaluating fixation mechanisms without background interference.
Additionally, we evaluated our model on four established FGVC benchmarks with increasing background complexity: CUB200-2011 \cite{CUB200} (12k images, 200 bird species), NABirds \cite{NABirds} (49k images, 555 species), Stanford-Dogs \cite{Dogs} (21k images, 120 breeds) and Food101 \cite{Food} (101k images, 101 categories). 
These datasets progress from moderate natural backgrounds to highly complex real-world imaging conditions.

\begin{wraptable}{r}{0.55\textwidth}
\centering
\vspace{-0.3cm}
\caption{Avg. top-1 test acc. on Stanford-Dogs. With $^\dagger$ being PPS trained with $N_{\text{train}}=4$.}
\label{tab:fgvc-dogs}
\setlength{\tabcolsep}{4pt}
\small
\begin{tabular}{@{}clccc@{}}
\toprule
 & Method & Backbone & Pretr. & Acc@1 \\
\midrule
\multirow{2}{*}{\rotatebox[origin=c]{90}{224\,px}}
  & RA-CNN \cite{RACNN} & VGG-16 & Im1k & 87.3 \\
  & MPSA \cite{MPSA} & SwinB & Im22k & 94.1 \\
  & PPS-MPSA (ours) & SwinB & Im22k & \textbf{94.2} \\
\cmidrule(l){2-3}
\multirow{3}{*}{\rotatebox[origin=c]{90}{384\,px}}
  & MPSA \cite{MPSA} & SwinB & Im22k & 94.7 \\
  & PPS-MPSA (ours) & SwinB & Im22k & 94.9 \\
  & PPS-MPSA$^\dagger$ (ours) & SwinB & Im22k & \textbf{95.1} \\
\cmidrule(l){2-3}
\multirow{3}{*}{\rotatebox[origin=c]{90}{448\,px}}
  & TransFG \cite{TransFG} & ViT-B16 & Im1k & 92.9 \\
  & FAL-ViT \cite{FAL-ViT} & ViT-B16 & Im1k & 91.1 \\
  & ACC-ViT \cite{ACC-ViT} & ViT-B16 & Im1k & 92.3 \\
\bottomrule
\end{tabular}
\end{wraptable}

\subsection{Fine-Grained Visual Classification}
\label{sec:fgvc}

We found consistent improvement over the vanilla backbones with only very few exceptions.
The margin is largest where small targets sit in cluttered scenes and where high intra-class variability rewards instance-adaptive attention, and smallest where the object already fills a clean, tightly cropped frame.
\noindent
On the moth benchmarks PPS-MPSA leads all baselines at 224px and at 384px with one exception (\Cref{tab:moths}). 
On trails on the near-saturated AMI-CA with little remaining headroom and results remain withing run-to-run noise.

\begin{wraptable}{r}{0.55\textwidth}
\centering
\vspace{-0.4cm}
\caption{Avg. top-1 test acc. on Food-101.}
\label{tab:fgvc-food}
\setlength{\tabcolsep}{4pt}
\small
\begin{tabular}{@{}lcccc@{}}
\toprule
Method & Backbone & Pretr. & Res. & Acc@1 \\
\midrule
FAL-ViT \cite{FAL-ViT} & ViT-B16 & Im1k & 448 & 91.8 \\
MPSA \cite{MPSA} & SwinB & Im22k & 224 & 93.7 \\
PPS-MPSA (ours) & SwinB & Im22k & 224 & \textbf{94.0} \\
\bottomrule
\end{tabular}
\end{wraptable}

\noindent
Two observations support our hypothesis. 
ECU backgrounds contain handwritten notes that invite overfitting, and the glimpses supply an additional self-augmentation path that reduces it in our experiments. 
Accuracy also rises on the noisier AMI subsets, indicating robustness to background clutter. 
Glimpses on ECU caused in 12.4\% a positive class flip and in 7.3\% a negative one.
Examples are shown in \Cref{fig:examples_nabirds_ecu}, where $\Delta p$ denotes the shift in the predicted class probability induced by the glimpses $\Delta p = p(y \mid \mathbf{z}) - p(y \mid \hat{\mathbf{z}}_L)$.
A positive value indicates the glimpses reinforced the model's decision.
Repeating the EU experiment with PPS-SwinB over three seeds left the accuracy unchanged to the first decimal (std below 0.1), indicating the reported margins are stable rather than seed artifacts.
CUB-200 presents tightly cropped birds with moderate clutter and little noise, so a downsampled glance already exposes the discriminative features and the gain from glimpses is small. 
We sample glimpses from 512px images, beyond which no further detail is available, and narrowly lead the best methods at 224 and 384px (\cref{tab:fgvc-cub}). 
NABirds places small bird regions in complex, noisy scenes with large scale and pose variation. 
Results are shown in \Cref{tab:fgvc-nabirds}, where the margin increased to $0.2$ accuracy points for MPSA using 224px inputs.
PPS suppresses background interference and recover fine detail when the target occupies a small fraction of the image.
On Stanford Dogs and Food-101 the advantage instead stems from high intra-class variability. 
PPS-MPSA improves over vanilla MPSA by 0.3\% on Food-101 and 0.4\% with four glimpses on Dogs.
The diverse sources, poses, lighting, and presentations in these sets reward an attention mechanism that adapts per instance and reads local geometry and texture.
The ViT based MPP methods trail at 448px despite their larger inputs. 


\begin{wraptable}{R}{0.5\textwidth}
\centering
\small
\vspace{-0.4cm}
\setlength{\tabcolsep}{6pt}
\caption{Avg. top-1 test accuracy on CUB with a priority map and loss ablation.}
\label{tab:ablation_prio_res_loss}
\begin{tabular}{@{}clccc@{}}
\toprule
& Method & SwinB & MPSA \\
\midrule
\multirow{2}{*}{$\mathbf{P}$} & upsampled & 90.4 & 91.4 \\
& native (ours)   & \textbf{90.7} & \textbf{91.5} & \\
\midrule
\multirow{3}{*}{$\mathcal{L}$} & REINFORCE \cite{Saccader} & 90.5 & 91.4 \\
& only $\mathcal{L}_{\text{task}}$ (Gumbel) & 90.6 & 91.4 \\
& \Cref{eq:sac_loss} (ours) & \textbf{90.7} & \textbf{91.5} \\
\bottomrule
\end{tabular}
\end{wraptable}

\subsection{Ablation Study}
\label{sec:ablation}

\textbf{Priority Map.}
We sample on $\mathbf{P}$ at its native feature resolution $H_P \times W_P$ rather than upsampling it to the image grid.
\Cref{tab:ablation_prio_res_loss} contrasts the two.
Native sampling exceeds the upsampled variant on both backbones at strictly lower cost.
Upsampling diffuses the gradient across pixels and adds compute with no accuracy return.
The coarse grid keeps the signal concentrated on each cell of $\mathbf{F}^{(S)}$ and shrinks the discrete location set, lowering sampler variance.
Thus, pixel-precise localization is unnecessary for FGVC.


\begin{wraptable}{R}{0.55\textwidth}
\centering
\vspace{-0.4cm}
\caption{Avg. top-1 test accuracy on CUB using a SwinB with and without MPSA.
The sampler's input feature map and the sampler itself are varied.}
\label{tab:ablation_sampling}
\setlength{\tabcolsep}{4pt}
\small
\begin{tabular}{@{}lccccc@{}}
\toprule
\multirow{2}{*}{Sampling} & \multirow{2}{*}{$\mathcal{F}$} & \multicolumn{2}{c}{$\mathbf{o} \sim \mathbf{F}$} & \multicolumn{2}{c}{$\mathbf{o} \sim \mathbf{P}$} \\
& & SwinB & MPSA & SwinB & MPSA \\
\midrule
Random & last & 90.3 & 91.4 & \textbf{90.8} & 91.3 \\
TopK & last & 90.2 & 91.4 & 90.5 & 91.3 \\
\midrule
NMS & full & 90.2 & 91.3 & 90.4 & \textbf{91.4} \\
NMS & mask & \textbf{90.6} & 91.4 & 90.3 & 91.3 \\
NMS & last & 90.0 & \textbf{91.5} & 90.7 & \textbf{91.4} \\
\bottomrule
\end{tabular}
\end{wraptable}

\noindent
\textbf{Sampling.}
\Cref{tab:ablation_sampling} varies the sampler and its input map. Two findings stand out.
First, the sampling source matters more than the sampler.
On a plain SwinB, drawing locations from the learned priority map $\mathbf{P}$ beats drawing them from the raw feature map $\mathbf{F}$ for every sampler.
MPSA shows the reverse: $\mathbf{F}$ already suffices (91.5) and $\mathbf{P}$ adds little (91.4).
The reason is that MPSA's CAM-condensed map \cite{GradCAM, MPSA} is itself a strong saliency signal, so sampling on it needs no extra head.

\begin{wraptable}{R}{0.2\textwidth}
\centering
\small
\vspace{-0.4cm}
\setlength{\tabcolsep}{6pt}
\caption{Avg. top-1 test acc. for MPSA with varying glimpses.}
\label{tab:num_glimpses}
\begin{tabular}{@{}lcc@{}}
\toprule
$N_{\text{test}}$ & EU & CUB \\
\midrule
2 & 98.2 & 91.4 \\
4 & 98.2 & 91.5 \\
8 & 98.2 & 91.5 \\
\bottomrule
\end{tabular}
\end{wraptable}

The priority head supplies the same signal to any backbone that lacks one.
Second, the edge mask of \citet{GFNet} is redundant under $\mathbf{P}$.
On $\mathbf{F}$ the mask helps (90.6 vs.\ 90.0), as raw Swin activations accumulate at borders.
On $\mathbf{P}$ it hurts (90.3 vs.\ 90.7): the learned head already avoids border accumulation, so masking only discards valid cells.
This confirms the claim of \Cref{sec:priomap_construction}.
The sampler itself is secondary.
Random, TopK, and NMS lie within $0.3$ points, so the differences sit at run-to-run noise on saturated CUB.
Random is a strong baseline, consistent with random-crop augmentation \cite{TIPooling}.
We also varied the number of test-time glimpses with results in \Cref{tab:num_glimpses}.
While the performance on EU is consistent ($2$ glimpses suffice for the well-aligned dataset), the performance increased with more glimpses on CUB. 
Similarly, we only found for Stanford Dogs an increase in performance when raising $N_{\text{train}}$ see \Cref{tab:fgvc-dogs}.

\noindent
\textbf{Fusion and Augmentation.}
We isolate sampling against augmentation and ablate our fusion. Averaging is the standard aggregator for sampled patches. 
\citet{DPS} report that mean pooling of patch embeddings beats concatenation on CUB, attributing the gap to reduced classifier overfitting. 
We adopt this averaging as the fusion baseline. 
\Cref{tab:augmentation} compares our sampler against training-time data augmentation (DA) and test-time augmentation (TTA) \cite{TIPooling}, both of which crop randomly and average the resulting signals.
Our sampler reaches the highest accuracy, which we argue stems from the contextual weighting in \Cref{eq:per_fix_fusion} (informed rather than random crops).

\begin{wraptable}{R}{0.55\textwidth}
\centering
\vspace{-0.4cm}
\setlength{\tabcolsep}{4pt}
\small
\caption{Avg. top-1 test accuracies on EU between our PPS, (training-time) data augmentation (DA), test-time augmentation (TTA), and a combination of both. DA and TTA use random sampling of target object patches and signal averaging.}
\label{tab:augmentation}
\begin{tabular}{@{}lcccc|l@{}}
\toprule
Method & Sampler & Fusion & $N_{\text{train}}$ & $N_{\text{test}}$ & MPSA \\ 
\midrule
Vanilla & - & - & 0 & 0 & 96.0 \\
DA & Rnd & Avg & 4 & 0 & 97.1 \\
TTA & Rnd & Avg & 0 & 4 & 96.0 \\
DA+TTA & Rnd & Avg & 4 & 4 & 96.4 \\
PPS (ours) & NMS & Attn & 4 & 4 & \textbf{98.0} \\
\bottomrule
\end{tabular}%
\end{wraptable}

\noindent
\textbf{Loss.}
\Cref{tab:ablation_prio_res_loss} compares three ways to train the priority map.
REINFORCE \cite{Saccader} treats sampling as a policy and optimizes it by policy gradient.
It is weakest (90.5), consistent with the high variance and baseline sensitivity of such estimators.
Replacing the \texttt{argmax} with a Gumbel softmax makes the sampler differentiable, so $\mathcal{L}_{\text{task}}$ reaches $\mathbf{P}$ directly and no auxiliary term is needed.
This removes the policy gradient but couples $\mathbf{P}$ to the task loss, inviting the encoder to reshape features for easy saliency rather than informative glimpses; it reaches 90.6.
Our decoupled objective (\Cref{eq:sac_loss}) supervises $\mathbf{P}$ with measured leave-one-out utility under a stop-gradient at $\mathbf{F}$, so the priority head and the backbone never compete.
It is best on both backbones (90.7, 91.5).
MPSA is near insensitive to this choice (91.4--91.5): its CAM features already localize, so how $\mathbf{P}$ is trained barely registers, and the loss matters mainly for a plain backbone.
Margins are small on saturated CUB, but the ranking is consistent and our loss avoids both the variance of RL and the shortcut risk of end-to-end differentiation.

\begin{wrapfigure}{R!}{0.5\linewidth}
    \vspace{-0.8cm}
  \begin{center}
    \includegraphics[width=0.95\linewidth]{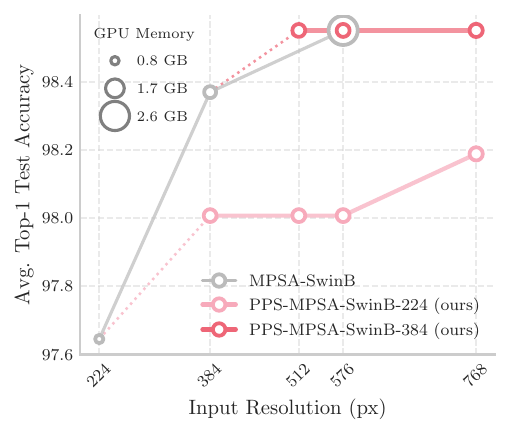}
  \end{center}
  \vspace{-0.5cm}
  \caption{
  Performance and peak GPU memory over input resolutions on EU.
  }
  \label{fig:res_acc_mem}
\end{wrapfigure}

\subsection{Complexity}
\label{sec:complexity}

A forward pass comprises the glance encoding $\phi(\mathbf{x}_L)$, $N$ glimpse encodings $\phi(\mathbf{x}_G^n)$, the location sampling from $\mathbf{P}$, and the cross-attention fusion.
From these components, the encoding by the backbone dominates.
Sampling and fusion are linear in $N$ but act on $D$ dimensional embeddings rather than the full feature tensor, so their cost is negligible against a backbone pass.
The compute is therefore $(1{+}N)$ backbone passes, linear in the glimpse budget $N$, and $N{=}0$ recovers the plain backbone exactly.
The $N{+}1$ encodings carry no inter-glimpse dependency and issue as a single batched call, which we exploit for throughput.
The serialized glimpse policies of \cite{RAM, GFNet} forbid this batching, since each glimpse conditions on the previous one.
Relative to a single high resolution pass we trade fewer pixels per pass for more passes, which are flexible to choose compared to \cite{RACNN, GFNet, DPS}.
On an A40 with a SwinB backbone the baseline reads a $384$\,px and $512$\,px input in $0.19$\,s and $0.53$\,s, while PPS ($N=2$) processes a $512$\,px source image in $0.46$\,s, at $48.8$, $116.1$, and $146.4$\,GFLOPs, respectively.
PPS is therefore not the cheapest option at every resolution. 
The benefit is memory rather than walltime. 
Peak memory is set by the glance resolution and the budget $N$, so it stays flat while a dense pass grows super-linearly.
Methods that encode the full high resolution input for context \cite{RAM, RACNN, PartCNN} pay the dense pass our glance resolution avoids.
Every encoder pass runs at $H_L{\times}W_L$, whereas a single high resolution pass scales with the full pixel grid $H{\times}W$.
This decoupling is sharpest in memory.
No full resolution activation map is ever materialized, so the peak memory of PPS stays essentially flat as the canvas grows, while the baseline rises super-linearly with resolution (\Cref{fig:res_acc_mem}).
There is an intersection point (here at $576$\,px) of both curves.
Below it a single low resolution pass is lighter than our $N{+}1$ passes, above it PPS is the cheaper model and the gap widens monotonically.
At high resolution the saving is the central practical benefit, since it is exactly the regime where a dense backbone pass becomes memory bound or infeasible.
The effect is strongest for backbones whose activation footprint grows with the input grid, convolutional networks in particular, which are moreover often widened to exploit higher resolution.
The added parameters are fixed across resolutions but scale with the backbone embedding dimension $D$. 
The fusion projections are $D \times D$, so the count grows with wider backbones. 
For SwinB the priority head and fusion block add 18.4M parameters.
A dedicated localization network \cite{RACNN, DPS, NTSNet} or recurrent controller \cite{RAM} instead adds overhead that scales with the backbone, not a fixed shallow head.
PPS thus decouples the memory cost of recognition from input resolution rather than trading it for a larger model.

\section{Conclusion}
We introduce the Plug and Play Saccader (PPS), a priority-guided sampling framework that equips any vision backbone with the ability to exploit high resolution inputs at the cost of the glance resolution and a small glimpse budget. 
A shallow head turns the backbone's own spatial features into a priority map, from which we sample glimpse locations by sequential Gaussian non maximum suppression. 
The glimpses are cropped, encoded by the weight-tied backbone, and fused with the glance through a residual cross attention block that weights each glimpse per head and per instance. 
A decoupled auxiliary loss supervises the priority map with leave one out glimpse utility, avoiding both the variance of reinforcement learning and the shortcut risk of end-to-end differentiation.
The design removes three constraints of prior attention guided methods: it needs no separate localization network, it treats the glimpse count as a runtime hyperparameter rather than an architectural choice, and it decouples memory cost from input resolution so peak memory stays essentially flat as the canvas grows. 
Across nine FGVC benchmarks, PPS improves consistently over its backbones. 
The added cost is a shallow priority head and a single fusion block fixed across resolutions.

\noindent
\textbf{Limitations and Future Work.}
Several avenues merit further investigation. 
The localization and extraction of glimpses can be read through the lens of affine transformations, which opens a perspective in which canonicalization \cite{Bootstrapping, ITS, ITS2} integrates naturally into saccadic vision pipelines to improve robustness. 
The additional backbone passes dominate the compute, so low rank approximations of these passes are a promising route to further reduce cost. 
Finally, the priority map offers a natural basis for sequential refinement through information theoretic view selection \cite{Denzler2002}, where early glimpses inform later ones.

\newpage
\section*{Acknowledgements}
Johann Schmidt was funded by the ZIM (KK5540302MS4).

\bibliography{main}

\newpage
\appendix
\section{Additional Results}
\label{app:additional-results}

We extend beyond Swin-Base/Tiny and add ConvNeXtv2-Base/Tiny and ViT-B16, which are pretrained on ImageNet22k and finetuned on ImageNet1k at 384\,px.
All numbers below stem from fully-converged runs (\emph{all} 100 epochs), gaining $90.7\!\rightarrow\!90.9$ (SwinB+PPS) and $91.5\!\rightarrow\!91.6$ (MPSA+PPS) on CUB compared to the main paper.
Baselines are retrained identically, so all comparisons stay like-for-like.

\paragraph{Fusion Ablation.}
To test whether gains come from added parameters rather than the saccadic design, we match the parameter count of PPS by a ConvNet head on top of the ConvNeXtv2-Base, vary fusion depth, and untie the encoders.
The results are shown in \cref{app:tab:parameter}.
PPS matches ConvNeXtv2-Base on the saturated EU column and exceeds it on CUB, while the parameter-matched ConvNet head \emph{degrades} EU ($-0.9$) at an equal budget.
A second fusion layer costs $~10$ M \#P and loses $1.1$/$0.7$ CUB points on SwinB/ConvNeXtv2-Base, buying at most $+0.7$ EU on the saturated ConvNeXtv2-Base.

\paragraph{Untying glance and glimpse encoders.}
The results are shown in \cref{app:tab:parameter}.
Untying glance and glimpse encoders nearly doubles the number of parameters (187\,M), while also decreasing performance.
Thus, weight-tying is not only efficient but more accurate.
Added capacity alone does not buy accuracy, the saccadic mechanism does.
Using a smaller glimpse encoder (Swin-Tiny, short SwinT) results in 128.5\,M parameters and a larger performance drop on CUB but a slight gain on EU.
The $18.4$\,M reported in \cref{sec:complexity} refer to PPS-MPSA-SwinB.
The addition (priority head + fusion) is fixed across input resolutions but scales with the embedding dimension ($+7.4$\,M for ViT-B16, $\approx\!{+}13$\,M for SwinB/ConvNeXtv2-Base). 

\begin{table*}[t]
\centering
\caption{
Glimpse encoder and fusion ablation of PPS with a ConvNeXtv2-Base (CNB) and a SwinB backbone.
Compared are the number of parameters (\#P) in millions (M) and the avg. top-1 test accuracies on CUB and EU.
}
\label{app:tab:parameter}
\small
\begin{tabular}{clccc}
\toprule
& Variant & \#P (M) ($\downarrow$) & CUB ($\uparrow$) & EU ($\uparrow$) \\
\midrule
\multirow{3}{*}{\rotatebox{90}{CNB}}
& vanilla                          & 87.9 & 90.2 & 96.2 \\
& + ConvNet (\#P-matched)          & 100.6 & 90.3 & 95.3 \\
& + \textbf{PPS} (ours)            & 101.0 & \textbf{90.5} & 96.2 \\
& \quad w/ 2 fusion layers         & 111.5 & 89.8 & \textbf{96.9} \\
\midrule
\multirow{4}{*}{\rotatebox{90}{SwinB}}
& + \textbf{PPS} (ours)            & 100.8 & \textbf{90.9} & \textbf{97.1} \\
& \quad w/ 2 fusion layers         & 110.5 & 89.8 & 96.9 \\
& \quad w/ untied SwinB glimpse enc.  & 187.0 & 90.3 & 96.4 \\
& \quad w/ untied SwinT glimpse enc.  & 128.5 & 89.9 & 96.7 \\
\bottomrule
\end{tabular}
\end{table*}

\begin{table*}[t]
\centering
\caption{
Ablation of priority encoders.
Alongside average top-1 test accuracy we report \emph{peak-in-box} (PiB).
This is the fraction of images whose hottest priority cell falls inside the ground truth box (flatness-robust), using CUB meta data.
}
\label{app:tab:prio-enc}
\small
\begin{tabular}{llccc}
\toprule
Model & Prio Enc. & CUB ($\uparrow$) & EU ($\uparrow$) & PiB CUB ($\uparrow$) \\
\midrule
SwinB & \textbf{Ours} & \textbf{90.9} & \textbf{97.1} & \textbf{97.7\%} \\
SwinB & CAM       & 90.7 & \textbf{97.1} & 88.5\% \\
SwinB & Attention & 90.8 & 96.9 & 87.9\% \\
SwinB & Spectral  & 90.4 & 96.9 & 47.2\% \\
\bottomrule
\end{tabular}
\end{table*}

\paragraph{Priority Encoder Ablation.}
To quantify fixation quality we keep the full PPS pipeline fixed and replace only the priority head with: (i) CAM/Grad-CAM (predicted class, no label leakage), (ii) TransFG-style accumulated attention (global average-pooling token as surrogate query as Swin Transformer has no class token), (iii) spectral-residual saliency (training-free floor) \cite{Hou2007}, (iv) and channel-L2 feature-norm (the $o\!\sim\!F$ source of \cref{tab:ablation_sampling}, 93.3\% PiB).
This ablation isolates the priority encoder and with it the priority map generation.
Result are in \cref{app:tab:prio-enc}.
EU is flat across all sources (96.9--97.1) and vanilla SwinB already reaches 97.06, so EU is not where the head pays off.
Accuracy on tightly-cropped CUB is likewise saturated and cannot separate these sources, which is precisely why component ablations there read as marginal in general\footnote{
The exception is the unsaturated moth benchmark ECU in \cref{tab:moths}.
PPS-MPSA gains $+3.9$ (on ECU with 224\,px) and $+3.1$ (on ECU with 384\,px) over MPSA.
}
However, saturated accuracy is a property of the benchmark, not evidence that the components are inert.
Where the signal is resolvable the ranking is decisive.
On CUB our head localizes with accuracy gains and at highest PiB.
This supports the benefit of our learned priority map, however, we note the difference as marginal when it comes to performance differences. 

\paragraph{PPS can also decrease performance.}
Additional to our results we also test PPS with a ViT-B16 \cite{ViT} backbone.
PPS lifts ViT-B16 by $+2.5$ on EU and $+0.1$ on NAB but loses $-0.3$ on CUB, which might be due to limited tuning.

\begin{table*}[t]
\centering
\caption{
Performance stability under differently seeded runs.
}
\label{app:tab:std}
\small
\begin{tabular}{lccc}
\toprule
PPS & EU (3 seeds) & CUB (3 seeds) & NAB (3 seeds) \\
\midrule
SwinB & 97.1$\pm$0.05 & 90.9$\pm$0.03 & 90.5$\pm$0.08 \\
MPSA+SwinB & 98.2$\pm$0.08 & 91.6$\pm$0.1 & 90.4$\pm$0.07 \\
\bottomrule
\end{tabular}
\end{table*}

\paragraph{Second-order pooling.}
Second-order pooling replaces the first-order aggregation in \cref{eq:per_fix_fusion} by compact bilinear pooling (CBP). 
We use Tensor Sketch \cite{TensorSketch} (signed square-root $+\,\ell_2$).
Matrix power norm is inapplicable as the glance--glimpse outer product is rank one.
However, second-order pooling does not improve beyond our first-order aggregation.
On EU this gives $97.1\!\rightarrow\!96.2$ and on CUB $90.9\!\rightarrow\!90.2$ with $+1.3$\,M \#P.
CBP aggregates \emph{features} via channel co-occurrences, orthogonal to PPS's spatial \emph{selection}. 
CBP is still a promising extension and we will cite both works.

\begin{table*}[t]
\centering
\caption{
Glimpse crop size sweep.
}
\label{app:tab:crop-size}
\small
\begin{tabular}{llccc}
\toprule
Model & Data & $s=0.15$ & $s=0.25$ & $s=0.35$ \\
\midrule
SwinB+PPS & CUB & 90.4 & \textbf{90.9} & \textbf{90.8} \\
SwinB+PPS & NAB & 89.4 & \textbf{90.5} & 89.6 \\
\bottomrule
\end{tabular}
\end{table*}

\paragraph{Crop size and object scale correlation.}
We provide a a glimpse-size/object-scale study on CUB (larger objects) and NAB (higher size variance) over crop scales $s$.
The results in \cref{app:tab:crop-size} confirm our expectation.
CUB prefers larger crops, while NAB a middle ground.

\end{document}